\documentclass[a4paper]{article}

\usepackage{cite}
\usepackage{amsmath,amssymb,amsfonts}
\usepackage{algorithmic}
\usepackage{graphicx}
\usepackage{textcomp}
\usepackage{xcolor}
\usepackage{hyperref}
\usepackage{makecell}

\def\BibTeX{{\rm B\kern-.05em{\sc i\kern-.025em b}\kern-.08em
    T\kern-.1667em\lower.7ex\hbox{E}\kern-.125emX}}

\graphicspath{ {figures/} }

\title{Financial Management System for SMEs: Real-World Deployment of Accounts Receivable and Cash Flow Prediction}

\author{Bartłomiej Małkus$^1$ \and Szymon Bobek$^2$ \and Grzegorz J. Nalepa$^2$}

\date{
	$^1$ Doctoral School of Exact and Natural Sciences \\ Jagiellonian University \\
    Cracow, Poland \\ \texttt{bartlomiej.malkus@doctoral.uj.edu.pl} \\%
	$^2$ Faculty of Physics, Astronomy and Applied Computer Science, Institute of Applied Computer Science, and Jagiellonian Human-Centered AI Lab \\ Jagiellonian University \\
    Cracow, Poland \\ \texttt{\{szymon.bobek, grzegorz.j.nalepa\}@uj.edu.pl}\\[2ex]%
}

\begin{document}

\maketitle

\begin{abstract}
Small and Medium Enterprises (SMEs), particularly freelancers and early-stage businesses, face financial management challenges due to limited data, small customer bases, and irregular cash flows. We present a deployed financial prediction system that combines accounts receivable prediction with cash flow forecasting for SME settings. The system integrates a binary classifier for invoice payment delays with a modular cash flow forecasting model designed to operate under incomplete historical data. A prototype was implemented and integrated into Cluee's platform, demonstrating practical feasibility for real-world SME financial management.
\end{abstract}

\section{Introduction}

Small and Medium Enterprises (SMEs), particularly early-stage businesses,\linebreak micro-enterprises, and freelancers, face financial management challenges that differ significantly from those of large corporations. With limited customer bases and smaller operational scales, these businesses are particularly vulnerable to cash flow disruptions and payment delays~\cite{afrifa2018working, shamsudin2015impending, farrell2016buffer}. A single delayed invoice can represent a substantial portion of monthly revenue, making financial prediction not just beneficial but critical for survival.

Traditional financial forecasting tools are designed for large enterprises with extensive historical data, dedicated finance teams, and complex enterprise systems. SMEs operate under fundamentally different constraints: limited historical data, basic invoicing systems, incomplete record-keeping, and minimal resources for financial analysis~\cite{tawil_trends_challenges}. In practice, this means sparse customer histories, heterogeneous income sources, and irregular cash flow patterns. These constraints also limit the use of feature-rich methods commonly assumed in the literature, since many SMEs cannot provide long customer histories or detailed operational data. At the same time, prediction outputs must remain understandable and actionable enough to support direct financial decisions, for example by indicating which invoices are risky or which cash-flow components drive the forecast.

To address this challenge, we developed and deployed a financial management system in collaboration with Cluee\footnote{see \url{https://www.cluee.app} for the company webpage.}, a startup providing self-organization and smart-budgeting tools for freelancers. Our system combines two components: accounts receivable prediction to anticipate payment delays, and cash flow forecasting to estimate future financial positions. These tasks are closely linked in SME workflows, since delayed receivables directly affect short-term liquidity and therefore influence broader cash flow planning. The system supports both standalone accounts receivable prediction for individual invoice management and integrated cash flow forecasting that incorporates payment delay estimates for more comprehensive financial planning.

This paper presents the real-world deployment of the system, focusing on practical challenges, architectural decisions, and business-oriented design rather than purely algorithmic contributions. We demonstrate how to build useful financial prediction systems under realistic SME constraints, including limited data availability, incomplete information, heterogeneous financial flows, and the need for transparent outputs.

Key contributions of this work include:
\begin{enumerate}
    \item a deployed integrated system combining accounts receivable prediction and cash flow forecasting for SME constraints;
    \item practical solutions for handling limited, incomplete, and heterogeneous financial data in real business environments;
    \item deployment-based validation and lessons learned from building financial prediction systems for resource-constrained businesses.
\end{enumerate}

\section{Related Work}

Financial prediction for businesses has been extensively studied, but most existing work focuses on large enterprises or settings with substantially richer data than those available in SMEs. Research in accounts receivable prediction typically examines large datasets and feature-rich customer histories~\cite{appel2019optimize, zeng2008using, hu2015predicting}, while cash flow forecasting studies commonly use data from power companies, construction projects, or other larger-scale operations~\cite{chen2019cash, cheng2015cash, cheng2011evolutionary}.

Accounts receivable prediction approaches employ machine learning methods such as gradient boosted trees, random forests, and neural networks on feature-rich datasets~\cite{appel2019optimize, appel2020predicting}. Zeng et al.~\cite{zeng2008using} highlighted the importance of customer- and history-based features, while Hu~\cite{hu2015predicting} analyzed prediction settings with extensive customer cooperation histories. Such feature requirements exceed what typical SMEs can provide, where invoicing records are often limited to basic transactional information and short customer histories.

Cash flow forecasting methods include ARIMA, Prophet~\cite{taylor2018forecasting}, and neural networks such as LSTM and GRU~\cite{weytjens2021cash, chen2019cash}. Cheng et al. proposed adaptive support vector regression~\cite{cheng2015cash} and evolutionary fuzzy decision models~\cite{cheng2011evolutionary} for construction projects. These approaches generally assume longer financial histories and more regular patterns than those observed in small businesses and freelance work.

Limited research addresses SME-specific constraints directly. Weytjens et al.~\cite{weytjens2021cash} compared machine learning to traditional methods, but still relied on relatively large datasets. The gap between enterprise-oriented methods and SME needs remains substantial, particularly for deployed systems that must jointly handle accounts receivable prediction and cash flow forecasting under sparse, incomplete, and heterogeneous data. Our work addresses this gap with a practical integrated solution designed for resource-constrained environments.

\section{Deployed Solution}

\subsection{System Architecture}

Our deployed system consists of two interconnected prediction modules integrated through a unified data processing pipeline. The architecture prioritizes modularity and transparency to accommodate the diverse data patterns and understandability requirements identified in SME workflows.

The accounts receivable prediction module operates as a binary classifier, determining whether invoices will be paid within a grace period (7 days past due date) or experience significant delays. The cash flow forecasting module employs a multi-component approach, with separate sub-models for different income and expense categories that are then aggregated into unified predictions.

\begin{figure}[ht]
\centering
\includegraphics[width=0.8\linewidth]{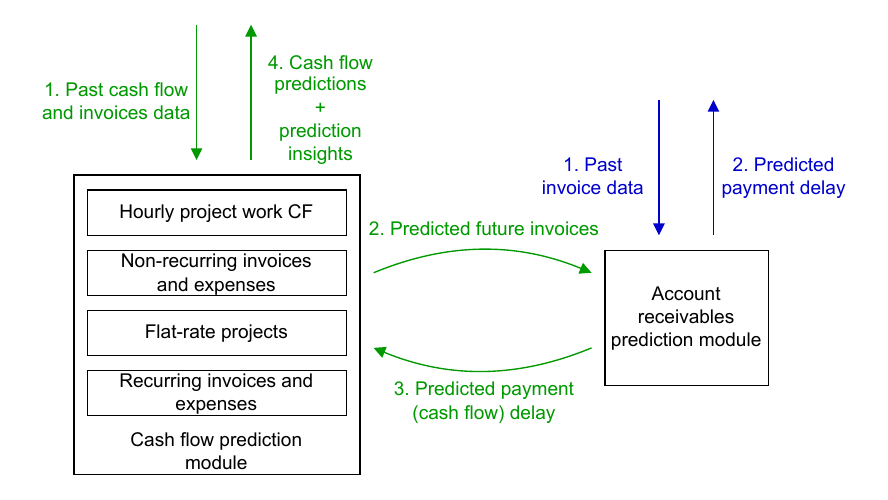}
\caption{System architecture supporting standalone accounts receivable prediction and integrated cash flow forecasting. The cash flow module consists of four sub-modules handling different income and expense categories.}
\label{fig:system_architecture}
\end{figure}

As illustrated in Figure \ref{fig:system_architecture}, the system supports standalone accounts receivable prediction and integrated cash flow forecasting that incorporates payment delay predictions. Raw invoice and project data is preprocessed and enriched with historical features, processed by the accounts receivable module, and then integrated into the cash flow forecasting pipeline, where payment delay predictions inform future cash flow estimates.

\subsection{Accounts Receivable Prediction Module}

The accounts receivable module addresses the challenge of predicting payment delays under limited customer history, which is common in SME settings. To maximize information extraction from sparse data, we designed a feature engineering approach combining basic invoice information (amount, payment terms, customer ID) with historical cooperation metrics derived from past interactions.

The engineered features include payment ratios (late payments versus total payments), average delay times, outstanding invoice counts, and monetary aggregations of past payment behavior. To capture changes in customer payment behavior over time, we also use moving-average-based trend features computed from recent payment delays and payment outcomes. Short-window averages capture recent changes, while longer-window averages reflect more stable historical patterns.

The module employs Support Vector Machines (SVM)~\cite{cortes1995support} as the primary classification algorithm, selected for their effectiveness on small datasets and clear decision boundaries. The model is trained separately for each business to account for differences in payment patterns across industries and customer relationships.

\subsection{Cash Flow Forecasting Module}

The cash flow forecasting module handles the heterogeneous nature of SME income and expense streams through a modular architecture with four specialized sub-modules. Each sub-module addresses a specific category of financial data, allowing the system to operate even when some data categories are missing or incomplete.

The hourly project work module calculates cash flows from ongoing projects based on recorded work sessions, hourly rates, and project timelines. Daily cash flows are computed as:

\begin{equation}
CF(d) = \sum_{p \in P} \sum_{t \in T_p} \frac{h_t}{d^{(w)}} \cdot w_t
\end{equation}

where $CF(d)$ is the predicted cash flow for one day, $P$ is the set of variable-rate projects, $T_p$ is the set of tasks in project $p$, $h_t$ is hours spent on task $t$, $w_t$ is the hourly wage, and $d^{(w)}$ is the number of working days in the historical period.

The non-recurring invoices and expenses module handles irregular payments using a conservative prediction approach:

\begin{equation}
CF(m_k) = \begin{cases}
\max(i, i_k) & \text{if } i > 0 \text{ and } i_k > 0 \\
i + i_k & \text{otherwise}
\end{cases}
\end{equation}

where $CF(m_k)$ is predicted cash flow for month $k$, $i$ is mean income from the last 6 months, and $i_k$ is planned income for month $k$. This conservative approach helps SMEs avoid overcommitting resources based on overly optimistic forecasts.

The flat-rate projects module manages fixed-fee projects by associating payment amounts with project completion dates, while the recurring invoices and expenses module handles subscription-based income and regular expenses with known schedules and amounts.

\subsection{Integration and Deployment}

The system operates in two modes to provide coherent financial projections. For standalone accounts receivable prediction, the system processes individual invoices to estimate payment delay likelihood. For integrated cash flow forecasting, the cash flow module directly calls the accounts receivable module to obtain payment delay predictions, which adjust timing assumptions in cash flow forecasts and improve short-term liquidity projections.

The system is implemented as a REST API using Flask and deployed on Google App Engine. This architecture supports integration with existing SME financial management tools while minimizing infrastructure requirements for the startup partner.

The API returns both compact summaries and component-level prediction details. For cash flow forecasting, the response includes contributions from each of the four sub-modules together with invoices flagged as potentially delayed. For accounts receivable prediction, the response includes the predicted delay outcome and recent customer payment-trend information.

\section{Real-World Performance and Impact}

\subsection{Evaluation Setup and Datasets}

Our evaluation focused on system effectiveness under realistic SME constraints. We used three datasets representing different scales of business operation: a startup-provided dataset from Cluee (297 invoices, 60 customers), and two public datasets filtered to match SME characteristics - IBM Late Payment Histories \cite{kaggle_ibm_dataset} (2,466 invoices) and Payment Date Dataset \cite{kaggle_generic_dataset} (12,071 invoices).

To better reflect typical SME customer relationships, we limited customer cooperation history to a maximum of 50 invoices per customer. We also applied a 7-day grace period for payment delay classification to account for payment processing delays common in small business transactions.

Because real cash flow data was limited, we developed a synthetic data generator producing realistic project-based work patterns for 1,000 simulated users over one year (422,306 work sessions). The generator incorporates variable working hours, multiple concurrent projects, and wage dynamics typical of freelance environments.

\subsection{Accounts Receivable Prediction Performance}

The accounts receivable module demonstrated consistent performance across datasets, with the startup dataset representing the most challenging scenario due to limited training data. We evaluated multiple classification algorithms using 5-fold cross-validation, with balanced accuracy \cite{brodersen2010balanced} as the primary metric to account for class imbalance.

Table \ref{tab:ar_performance} shows comparative performance across methods. Additional evaluation using F1-score showed consistent trends with balanced accuracy. Based on these results, we selected SVM as the primary algorithm due to its consistent performance across datasets. While gradient boosting methods such as XGBoost are commonly used for tabular data, our focus was on models that remain interpretable and robust under very limited customer histories, which motivated the use of simpler models in this setting. On the startup dataset, a balanced accuracy of 0.56, while modest, is meaningful given the severe data constraints, with 50\% of customers having only 1--2 invoices.

\begin{table}[ht]
\centering
\caption{Balanced accuracy of accounts receivable prediction across datasets.}
\label{tab:ar_performance}
\begin{tabular}{|l|c|c|c|}
\hline
\textbf{Method} & \textbf{Startup} & \textbf{IBM} & \textbf{Kaggle} \\
\hline
Decision Tree & 0.54 & 0.68 & \textbf{0.70} \\
Random Forest & 0.52 & 0.69 & 0.67 \\
SVM & \textbf{0.56} & \textbf{0.72} & 0.68 \\
kNN & \textbf{0.56} & 0.68 & 0.66 \\
Naive Bayes & 0.54 & \textbf{0.72} & 0.62 \\
\hline
\end{tabular}
\end{table}

Moving average features improved performance across datasets by 3--5 percentage points. The system also remained applicable in settings with very limited history, even with as few as two previous customer interactions.

\subsection{Cash Flow Forecasting Results}

Cash flow forecasting evaluation highlighted the importance of model simplicity and transparency in resource-constrained environments. Our custom modular approach consistently outperformed traditional time series methods (ARIMA, Prophet, SVR) when training data was limited.

In the most challenging scenario, predicting 11 months based on only 1 month of historical data, our approach achieved 11.85\% mean absolute percentage error (MAPE), compared to 159.40\% for Prophet and 166.24\% for SVR. This result illustrates the difficulty of applying standard time series approaches in SME cold-start settings.

Table \ref{tab:performance} summarizes performance across representative evaluation scenarios, showing consistent advantages of the proposed approach under SME-typical data constraints.

\begin{table}[ht]
\centering
\caption{MAPE (\%) across representative cash flow forecasting scenarios. CF (X/Y months) denotes forecasting Y months ahead using X months of training data.}
\label{tab:performance}
\begin{tabular}{|l|c|c|c|}
\hline
\textbf{Method} & \textbf{CF (9/3 months)} & \textbf{CF (6/6 mo)} & \textbf{CF (1/11 mo)} \\
\hline
SVM/SVR & 14.19 & 19.11 & 166.24 \\
Prophet & 14.29 & 19.01 & 159.40 \\
ARIMA & 19.61 & 10.72 & -$^{\mathrm{a}}$ \\
Our Method & \textbf{13.06} & \textbf{9.41} & \textbf{11.85} \\
\hline
\multicolumn{4}{l}{\makecell[l]{$^{\mathrm{a}}$ARIMA is missing, as there is not enough data in one month for\\AutoARIMA to determine the order. ARIMA is performed on weekly\\aggregated data as it performed much better than on the daily one.}} \\
\end{tabular}
\end{table}

The modular architecture was important for handling incomplete data categories. Unlike monolithic approaches that fail when data is missing, the system maintained useful predictions when only a subset of data sources was available.

\section{Discussion and Future Directions}

This work presented the design, implementation, and deployment of an integrated financial prediction system for SMEs operating under severe data and resource constraints. The system combines accounts receivable prediction with cash flow forecasting in a unified architecture, enabling short-term liquidity management and medium-term planning. The results indicate that useful prediction is possible even under sparse and incomplete data, provided that the methods are designed around SME constraints. The modular architecture was also important operationally, as it allowed the system to remain useful under heterogeneous data availability and partial financial information.

In practice, predictions were integrated into the platform as lightweight decision-support signals rather than prescriptive recommendations. Accounts receivable predictions identified invoices with elevated delay risk, while cash flow forecasts emphasized near-term liquidity trajectories adjusted for expected payment delays. This reflected feedback from the startup team that users preferred conservative, short-term insights and outputs whose reasoning could be inspected, such as which invoices were flagged as risky and which cash-flow components contributed most to the forecast.

The study remains limited by the small startup dataset and the use of synthetic data for part of the cash flow evaluation. Future work will focus on broader validation in more diverse SME settings, richer integration of receivable risk into forecasting, and extension of the system as more deployment data becomes available.

\bibliographystyle{IEEEtran}
\bibliography{bibliography}

\end{document}